# Empirical Mode Decomposition and Graph Transformation of the MSCI World Index: A Multiscale Topological Analysis for Graph Neural Network Modeling

Agustín M. de los Riscos [1], Julio E. Sandubete [1,*], Diego Carmona-Fernández [2], León Beleña [1]

[1] Faculty of Law, Business and Government, Universidad Francisco de Vitoria, 28223 Madrid, Spain.
[2] Electrical, Electronic and Control Engineering Department, Universidad de Extremadura, 06006 Badajoz, Spain
* Correspondence: je.sandubete@ufv.es

**Abstract:** This study applies Empirical Mode Decomposition (EMD) to the MSCI World index and converts the resulting intrinsic mode functions (IMFs) into graph representations to enable modeling with graph neural networks (GNNs). Using CEEMDAN, we extract nine IMFs spanning high-frequency fluctuations to long-term trends. Each IMF is transformed into a graph using four time-series-to-graph methods: natural visibility, horizontal visibility, recurrence, and transition graphs. Topological analysis shows clear scale-dependent structure: high-frequency IMFs yield dense, highly connected small-world graphs, whereas low-frequency IMFs produce sparser networks with longer characteristic path lengths. Visibility-based methods are more sensitive to amplitude variability and typically generate higher clustering, while recurrence graphs better preserve temporal dependencies. These results provide guidance for designing GNN architectures tailored to the structural properties of decomposed components, supporting more effective predictive modeling of financial time series.

## 1. Introduction

Analyzing financial time series poses specific challenges derived from the market, whose dynamics are typically neither linear nor stationary. Traditional time series decomposition methods, including wavelet transforms and Fourier analysis, are based on predefined basis functions that often prove inadequate for capturing the inherent properties of financial datasets. The Empirical Mode Decomposition (EMD), originally proposed by (8), has proven to be highly suitable for the analysis of financial time series, as it provides an adaptive and data-driven methodology that operates without making assumptions about the underlying data structure. Unlike traditional techniques, EMD adaptively locates local extrema and constructs envelopes to isolate oscillatory components, making it highly effective for decomposing financial series. Each of the resulting series is characterized by different time-dependent amplitude and frequency values.

The effectiveness of EMD in the analysis of financial time series stems from this technique's ability to adapt to the fundamental properties that define stock market data: non-stationarity resulting from prolonged trends, nonlinear dynamics evident through volatility clustering and regime transitions, and variable amplitude and frequency patterns according to different market states. These characteristics make financial time series particularly suitable for EMD analysis, given that the technique is explicitly designed to isolate intrinsic mode functions (IMFs) from signals exhibiting these characteristics. The MSCI World index, which serves as a global market indicator, fulfills the properties enumerated above, positioning it as an appropriate case study for applying EMD to the decomposition of time series in this sector.

Although EMD provides a robust decomposition framework, the resulting IMFs require advanced modeling techniques to represent structural and temporal relationships. To achieve this objective, this line of work proposes graph neural networks (GNNs), given their ability to effectively learn from graph-structured data and identify both localized and global patterns through message- passing mechanisms. However, to utilize GNNs in the analysis of IMFs, it is essential to represent temporal sequences as nodes and edges through data transformations, while always maintaining the temporal order of the series.

To establish the foundations for GNN-based modeling, this article focuses on the application of EMD techniques and the transformation of components into graphs using various representation methods. At a more detailed level, the scope of this research is defined to include: validating the applicability conditions of EMD for the MSCI World index through statistical tests; decomposing the index into more than one component using EMD techniques, with a comprehensive characterization of the energy, variance, frequency, and amplitude attributes of each component; converting all IMFs into graph representations using existing algorithms; and calculating metrics of the obtained graphs (clustering coefficients, centrality measures, connectivity patterns) that illustrate the structural characteristics of each decomposed component. The execution of all these steps is accompanied by an interpretation and discussion of the results obtained.

The remainder of the article is organized into four main sections. Section 2 develops the theoretical and methodological foundations of the study, where the MSCI World index dataset is described, the theory of Empirical Mode Decomposition (EMD) along with its variants is presented, the prior considerations necessary for its effective application are discussed, and the methods for transforming time series into graphs through natural visibility (NVG), horizontal visibility (HVG), and recurrence algorithms are introduced. Subsequently, Section 3 presents the empirical findings, which are structured into three blocks: on one hand, the validation of the appropriate conditions for EMD application (see Section 3.1); on the other, the EMD decomposition of the MSCI World index along with the characterization of the obtained intrinsic mode functions (IMFs); and finally, the analysis of graph transformations, addressing their structural properties and the calculated metrics. To conclude, Section 4 offers a synthesis of the main findings and proposes directions for future research.

## 2. Methodology

### 2.1 Description of the data

The analysis is based on the daily closing prices of the MSCI World index, see Figure 1. The analyzed dataset spans from January 12, 2012, to November 26, 2025, and comprises 3490 daily observations. This period covers various market regimes, including bull markets, corrections, and periods of high volatility. The price series ranges from a minimum of 38.04$ in US dollars (June 5, 2012) to a maximum of 186.64$ (October 28, 2025), representing a cumulative return of 376.27 % during the analysis period. Daily returns are calculated as percentage variations, with a mean daily return of 0.0506 % and a standard deviation of 1.0780 %. The 30-day moving average volatility is 0.9417 %, with a maximum value of 5.1228 % during periods of high volatility.

### 2.2 Empirical Mode Decomposition. Theory and variants

The EMD decomposition, introduced by (8), is an adaptive data-driven method for decomposing non-stationary and nonlinear time series into a finite set of oscillatory components called intrinsic mode functions (IMF). Unlike traditional decomposition methods that rely on predefined basis functions (for example, Fourier transforms that use sinusoidal bases or wavelet transforms that use predefined mother wavelets), EMD adaptively identifies local extrema and constructs envelopes to extract oscillatory components at different temporal scales.

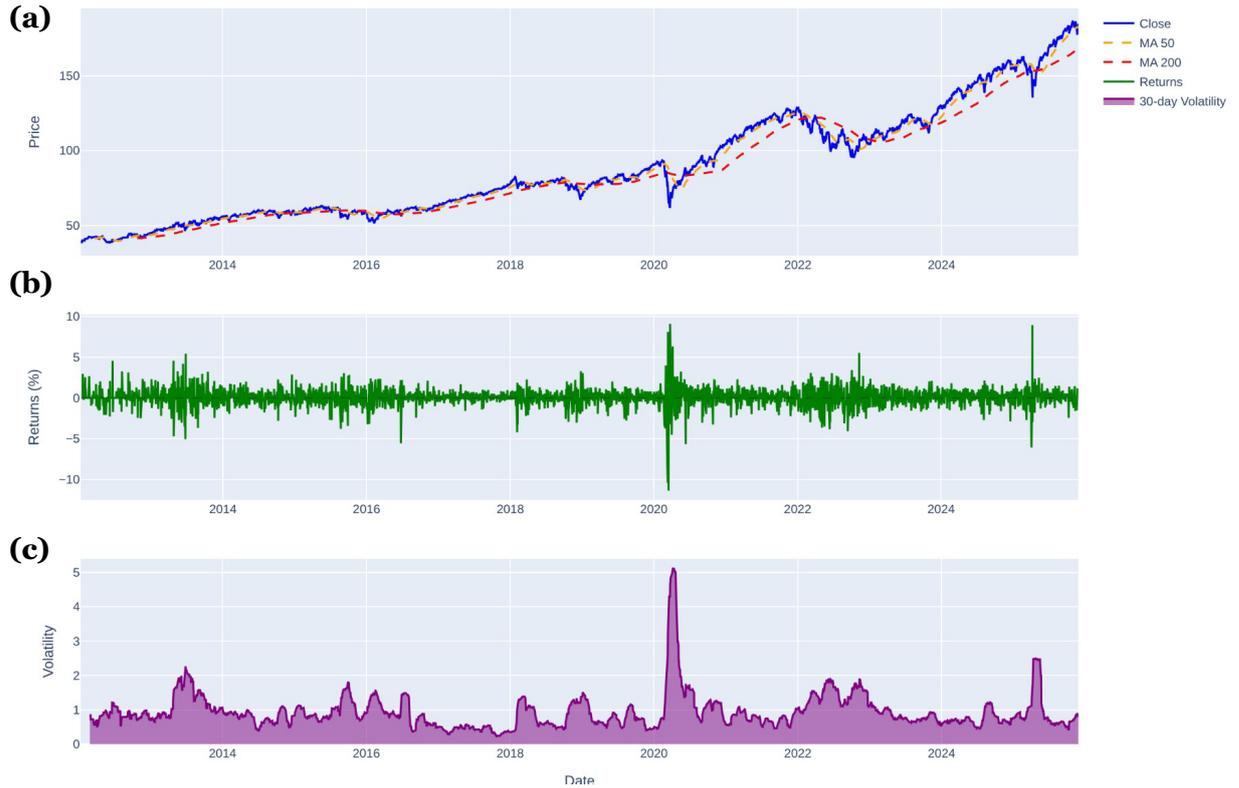

**Figure 1.** Analyzed data of the MSCI World. a) Time series trends of closing prices for the MSCI World index from January 2012 to November 2025, accompanied by the 50 and 200-day moving averages that allow identifying short and long-term patterns. b) Daily returns calculated as percentage variations, which capture the daily variability of the index. c) 30-day moving volatility, which measures the degree of variation of returns in a 30-day moving window. This metric allows identifying periods of higher risk and market stability, showing pronounced peaks during market events such as those observed in 2020.

The fundamental principle of EMD is the sifting process, which iteratively extracts IMFs by identifying local maxima and minima, constructing upper and lower envelopes through cubic spline interpolation, calculating the mean envelope, and subtracting it from the signal. This process continues until the resulting signal satisfies the IMF conditions: the number of extrema and zero crossings must differ by at most one, and the mean value of the upper and lower envelopes must be zero at any point. This process ends when a stopping criterion is met, typically based on a standard deviation threshold between consecutive sifting iterations.

However, standard EMD has several limitations, among which the mode mixing problem stands out. When decomposing with the original EMD method, oscillations of different scales tend to mix into a single IMF and similar oscillations may be split across several IMFs. This problem is magnified in the presence of intermittent signals and noise, which can make the decomposition unstable and non-unique.

To address these limitations, several EMD variants have been developed. The Ensemble Empirical Mode Decomposition (EEMD) (16) mitigates mode mixing by adding white noise to the signal multiple times, performing EMD on each noise-perturbed version, and averaging the resulting IMFs. The ensemble average cancels out the noise and preserves the real components of the signal, resulting in more stable and physically meaningful decompositions. EEMD requires two key parameters: the number of ensemble trials and the noise amplitude, which is usually expressed as a fraction of the signal's standard deviation.

In subsequent work, the Complete Ensemble Empirical Mode Decomposition with Adaptive Noise (CEEMDAN) (15) improves upon EEMD by adding noise adaptively at each stage of the decomposition, rather than only at the initial stage. This approach reduces the number of ensemble trials required, while maintaining the quality of the decomposition and further reducing residual noise in the IMFs.

To obtain good results in the decomposition of the MSCI World, it is enough to select the EEMD variant (see Section 3) with the execution parameters defined in Table 1.

**Table 1.** EEMD decomposition parameters

| Parameter | Value |
| --- | --- |
| Maximum number of IMFs | 14 |
| Standard deviation threshold (SD_thresh) | 0.25 |
| Minimum sifting iterations (S_number) | 8 |
| Fixed iterations (FIXE_H) | 5 |
| Ensemble trials | 100 |
| Noise width | 0.05 (5% of signal's std dev) |

**2.3 Prerequisites for effective EMD application**

For EMD analysis to be effective, the time series must satisfy the following key conditions: *(i)* Non-stationarity. EMD decomposition offers optimal performance with non-stationary series, as it can adaptively capture trends and components at different temporal scales. Therefore, the stationarity of the studied MSCI World time interval is checked using the Augmented Dickey-Fuller (ADF) test (4) and the Kwiatkowski-Phillips-Schmidt-Shin (KPSS) test; *(ii)* Non-linearity. Applying EMD techniques to the series does not require linearity assumptions. Non-linearity is evaluated through multiple approaches: using the Brock-Dechert-Scheinkman (BDS) test (2) on the residuals of an AR(1) model, through autocorrelation analysis of squared returns to detect ARCH/GARCH effects (6; 1), and finally using the Ljung-Box test (11) on the squared returns of the index; *(iii)* Variable amplitude and frequency: EMD excels at capturing modes with time-varying amplitude and frequency. Therefore, amplitude variability is analyzed through continuous volatility measures (252-day windows corresponding to one trading year) and frequency variability through zero-crossing analysis. The results of all tests listed in this section are described in Section 3.1.

**2.4 Time series to graph transformations**

The transformation of time series into graph representations allows us to analyze the structural properties of temporal data using graph theory techniques and network analysis. Furthermore, converting to graphs enables the application of graph neural networks and offers an alternative perspective for understanding the underlying dynamics of decomposed time series components. In this study, we employ three complementary graph transformation methods to convert the extracted IMFs into graph structures: the natural visibility graph (NVG), the horizontal visibility graph (HVG), and recurrence-based networks.

**2.4.1 Visibility algorithms**

Visibility graph algorithms, introduced by (10), transform a time series into a graph by establishing connections between data points based on geometric visibility criteria. The fundamental principle is that two points in the time series are connected if no intermediate point obstructs the line of sight between them.

The variant called the natural visibility graph (NVG) algorithm constructs a graph in which each data point (ti, yi) of the time series becomes a node, and an edge exists between nodes i and j if, for all intermediate points k such that i < k < j, the following condition is satisfied:

$$y_k < y_i + (y_j - y_i)\frac{t_k - t_i}{t_j - t_i} \quad (1)$$

This geometric criterion ensures that the line segment connecting points i and j lies above all intermediate data points, establishing a natural visibility relationship. NVG preserves important properties of the original time series, such as periodicity, fractality, and hierarchical structure, making it suitable for analyzing complex temporal patterns in financial data.

In a second variant, the horizontal visibility graph (HVG) algorithm (12) simplifies NVG by imposing a stricter horizontal visibility constraint. Two nodes i and j are connected if, for all intermediate points k with i < k < j, the following condition is satisfied:

$$y_k < \min(y_i, y_j) \quad (2)$$

These two transformations have demonstrated utility in various recent applications in the financial domain. An example is found in Han et al. (2022), where they developed MOHVGCA (Multiscale Online-Horizontal-Visibility-Graph Correlation Analysis), a method that employs HVG to quantify scale-dependent correlations between financial time series, revealing dynamic associations between WTI oil futures and various currencies during oil crises (7). More recently, Chen (2025) proposed the use of rising visibility graphs to extract topological motifs from trend structures in stock indices, demonstrating that these recurrent topological structures can be identified and used to forecast emerging patterns in financial time series (3). These studies demonstrate the ability of visibility algorithms to reveal hidden structural characteristics in time series from various domains, validating their application in the analysis of complex financial data as addressed in this work.

**2.4.2 Recurrence-based algorithms**

As an alternative to visibility algorithms, the recurrence matrix can be employed as an adjacency matrix to transform time series into graph representations by identifying recurrence patterns within the dynamical system. This method captures the recurrence structure of the series in a higher-dimensional embedding space. This effectively reveals both the geometry of the underlying attractor and the essential dynamical properties of the system.

The construction of recurrence graphs begins with the determination of optimal embedding parameters using geometric and information-theoretic criteria. The delay parameter $\tau$ is selected by analyzing mutual information, specifically identifying the first minimum of the mutual information function between the original series and its time-delayed version. This ensures that the delayed coordinates provide maximum information about the system state while minimizing redundancy. Subsequently, the optimal embedding dimension d is estimated using the false nearest neighbors (FNN) method. This technique identifies the minimum dimension required to unfold the attractor by detecting false neighbors that appear close in low dimensions but are separated in higher dimensions. The FNN method systematically increases the embedding dimension until the proportion of false neighbors falls below a threshold, typically 10%.

Once $\tau$ (time delay) and d (embedding dimension) are determined, the embedding space is constructed using delay coordinate embedding. Each point Xi in this space is defined as a d-dimensional vector:

$$X_i = [Y_i, Y_{i+\tau}, Y_{i+2\tau}, \ldots, Y_{i+(d-1)\tau}] \quad (3)$$

where Xi represents the i-th point in the embedding space and Yi denotes the value of the original time series at time i. The resulting recurrence graph preserves the topological and geometric properties of the underlying attractor, enabling the analysis of dynamical invariants such as correlation dimension, Lyapunov exponents, and entropy measures. Node features in the recurrence graph are assigned using the first component of the embedding vectors, maintaining a connection with the original time series values.

From the generated embedding space, the recurrence graph is constructed by computing pairwise distances between all embedding vectors. Two nodes i and j in the graph are connected if the distance between their corresponding embedding vectors Xi and Xj is below a threshold ε. This is formally expressed through the (i, j) element of the recurrence matrix R:

$$R_{ij} = \begin{cases} 1 & \text{if } \|X_i - X_j\| \leq \varepsilon \\ 0 & \text{otherwise} \end{cases} \quad (4)$$

where Rij is the (i, j)-th element of the recurrence matrix. The threshold ε is typically chosen as a percentile of the pairwise distance distribution, ensuring that the graph captures the most significant recurrent structures while maintaining computational tractability.

Numerous studies have demonstrated that the transformation of time series to recurrence graphs has proven particularly useful in financial market analysis, where capturing nonlinear dynamics and complex patterns is crucial for prediction and anomaly detection. In this context, Su et al. (2025) (14) developed a hybrid dual-branch model that integrates recurrence plots with transposed transformers for stock trend prediction. In their approach, closing price time series are transformed into recurrence plots using sliding windows of historical segments.

The first branch of the model generates RP images from these univariate sequences and extracts recurrence quantification analysis (RQA) measures, which capture nonlinear characteristics and temporal relationships between points. These RQA measures include metrics such as recurrence rate (RR), determinism (DET), and laminarity (LAM), which quantify the predictability and structure of the dynamical system. The second branch processes multivariate time series features using a transposed transformer, and both branches are merged to classify the next-day trend, either up or down. The authors evaluated the model on seven selected stocks and demonstrated that combining recurrence-based features with sequence modeling through transformers produces superior accuracy and F1-scores compared to traditional machine learning and deep learning methods. This approach illustrates how recurrence plots can capture nonlinear dynamics and relationships between temporal points that are difficult to detect using conventional time series analysis methods.

On the other hand, Song and Li (2024) (13) proposed the use of multiplex recurrence networks (MRNs) for detecting early warning signals of critical transitions and crashes in stock markets. In their methodology, multidimensional return vectors are transformed into multiplex recurrence networks, where each dimension, corresponding to different assets or indicators, forms a network layer. Recurrence relationships are established within each layer and between layers, capturing both the individual dynamics of each asset and the interdependencies between them. The authors calculate topological indicators of MRNs, specifically average mutual information and average edge overlap, to monitor changes in the recurrence structure over time. These indicators are evaluated both in simulated systems and empirical data from Chinese and US markets. The results demonstrate that indicators derived from MRNs can signal imminent state transitions in the system and provide insights into significant historical market events.

Following the review of transformation types, it is concluded that applying visibility and recurrence algorithms to IMF (Intrinsic Mode Function) components is of considerable interest. This methodological strategy enables a more comprehensive analysis of the resulting graph, as visibility complements information regarding the temporal structure of the sequence, while recurrence reveals the relationships of the dynamical system inherent to the time series. As discussed in Section 4, applying other types of transformations will be considered in future research lines.

## 3. Results

### 3.1 Suitable conditions for EMD application

A battery of tests is performed to confirm an adequate study via EMD decomposition of the indexed series. Our comprehensive evaluation of the MSCI World index confirms that the time series meets the three fundamental conditions for effective EMD analysis. The results of the three evaluated conditions are presented below.

### 3.1.1 Stationarity

Stationarity tests provide strong evidence of the non-stationarity of the price series, which is ideal for EMD. The Augmented Dickey-Fuller (ADF) test on the closing price series yields a test statistic of 1.0542 with a p-value of 0.9948, which does not allow rejecting the null hypothesis of a unit root at all conventional significance levels. The critical values at the 1%, 5% and 10% levels are -3.4322, -2.8624 and -2.5672, respectively, all substantially more negative than the test statistic. The Kwiatkowski-Phillips-Schmidt-Shin (KPSS) test, which has stationarity as the null hypothesis, produces a test statistic of 1.2110 with a p-value of 0.0100, which rejects the null hypothesis of stationarity at the 5% significance level. Conversely, the returns series with first difference is confirmed as stationary in both tests: ADF statistic of -19.5227 (p-value < 0.0001) and KPSS statistic of 0.0341 (p-value = 0.1000). These complementary results confirm that the price series exhibits non-stationary behavior driven by long-term trends, making it highly suitable for EMD decomposition.

### 3.1.2 Linearity

Multiple tests provide consistent evidence of strong nonlinear dynamics in the MSCI World index. The Brock-Dechert-Scheinkman (BDS) test, applied to the residuals of an AR(1) model, detects significant nonlinear dependencies across multiple embedding dimensions. In dimensions 3, 4 and 5, the BDS test statistics are 20.99, 24.95 and 27.83, respectively, all with p-values < 0.0001, which strongly rejects the null hypothesis of independence and indicates the presence of a nonlinear structure.

The distributional analysis further supports the nonlinear characterization. The price series shows significant deviations from normality, with a skewness of 0.7515 and a kurtosis of -0.3538. The Jarque-Bera test statistic yields a p-value < 0.0001, which strongly rejects the null hypothesis of normality. These findings collectively demonstrate that the MSCI World index exhibits rich nonlinear dynamics, making EMD an adequate decomposition method.

### 3.1.3 Volatility in amplitude and frequency

The autocorrelation analysis of squared returns reveals substantial volatility clustering, a distinctive feature of nonlinear ARCH/GARCH effects. We observe 20 significant autocorrelation lags out of the 20 analyzed, with a maximum autocorrelation of 0.9978, indicating strong volatility persistence. The Ljung-Box test on squared returns produces a test statistic of 34 210.69 with a p-value < 0.0001, which provides overwhelming evidence of ARCH/GARCH effects and confirms nonlinear dependence in the variance process.

**3.2 EMD Decomposition**

The decomposition of the MSCI World index was performed using the EEMD (Ensemble Empirical Mode Decomposition) variant with the parameters specified in Table 1. This methodological choice is based on EEMD's ability to mitigate the mode mixing problem through the addition of Gaussian white noise and the averaging of multiple decompositions, resulting in a more stable and physically meaningful decomposition.

The application of EEMD to the closing price series of the MSCI World index, which comprises 3490 daily observations from January 12, 2012, to November 26, 2025, produced a decomposition into 10 intrinsic mode functions (IMFs) and a monotonic residue. The verification of the residue confirms that it is strictly monotonic (without local extrema), thus fulfilling the fundamental termination criterion of the sifting process in EMD.

**3.2.1 Characterization of the IMFs**

Each extracted IMF represents oscillations at different temporal scales, from high-frequency components (IMF1) to low-frequency components (IMF10). Table 2 presents a comprehensive summary of the key metrics calculated for each IMF: energy, variance, dominant frequency (expressed in cycles over the complete period), average amplitude and standard deviation.

The analysis of the metrics reveals clear patterns in the structure of the IMFs. Lower-order IMFs (IMF1 to IMF4) exhibit high dominant frequencies (from 98 to 1421 cycles) and relatively small amplitudes (less than 0.87), indicating that they capture short-term oscillations and high- frequency noise. As the order of the IMF increases, a systematic decrease in dominant frequency and a progressive increase in variance and average amplitude are observed. Higher-order IMFs (IMF9 and IMF10) present dominant frequencies of 1 cycle and high average amplitudes (91.94), suggesting that they capture the long-term trend of the series. IMF10 shows the highest energy ($2.97 \times 10^7$) and variance (1058.69), indicating that it contains most of the variability of the original signal, with a contribution of 88.47% of the total energy.

**Table 2.** Main metrics of the IMFs extracted via EEMD.

| IMF | Energy | Variance | Frequency (cycles) | Amplitude Average |
|---|---|---|---|---|
| $IMF_1$ | $1.75 \times 10^3$ | 0.5024 | 1421.0 | 0.8659 |
| $IMF_2$ | $8.39 \times 10^2$ | 0.2405 | 552.0 | 0.5878 |
| $IMF_3$ | $8.74 \times 10^2$ | 0.2505 | 284.0 | 0.5718 |
| $IMF_4$ | $1.55 \times 10^3$ | 0.4430 | 98.0 | 0.7310 |
| $IMF_5$ | $3.27 \times 10^3$ | 0.9359 | 66.0 | 1.0165 |
| $IMF_6$ | $5.78 \times 10^3$ | 1.6523 | 27.0 | 1.3779 |
| $IMF_7$ | $1.31 \times 10^4$ | 3.7452 | 11.0 | 2.0538 |
| $IMF_8$ | $1.31 \times 10^5$ | 37.4858 | 4.0 | 6.5590 |
| $IMF_9$ | $3.71 \times 10^6$ | 153.9538 | 1.0 | 91.9395 |
| $IMF_{10}$ | $2.97 \times 10^7$ | 1058.6908 | 1.0 | 91.9395 |

**3.2.2 Validation of the decomposition**

The validation of the EEMD decomposition was performed by reconstructing the original signal by summing all IMFs. The reconstruction using only the IMFs (without including the residue) generates an RMSE of 30.49 and an MAE of 28.50, which demonstrates the importance of the residue for capturing the long-term trend and ensuring complete signal reconstruction.

The residual component, calculated as the difference between the original signal and the sum of all IMFs, presents a mean of -28.50, a standard deviation of 27.75, and a range from 47.01 to 132.48. Its monotonic nature confirms that the sifting process was completed correctly and that no additional oscillations remain to be extracted by the EEMD algorithm. Figure 2 visually shows the complete decomposition of the MSCI World index, where the progression from high-frequency IMFs (IMF1) to low-frequency IMFs (IMF10) and the residue can be observed, illustrating how each component captures oscillations at different temporal scales and how the sum of all of them reconstructs the original signal.

The observation that the sum of the IMFs (without including the residue) presents systematically higher values than the original signal, as evidenced in the analysis of the transformed graphs, is an expected and valid result within the EEMD decomposition framework. This behaviour arises from the iterative nature of the sifting process, where each IMF is extracted through the progressive elimination of mean envelopes, which can introduce small numerical accumulations that cause the sum of the IMFs to slightly exceed the original signal. In EEMD, the averaging of multiple ensemble realizations can amplify these discrepancies, as each individual realization may present deviations that, when averaged, do not completely cancel out. The negative residue (mean of -28.50) acts as a correction term that compensates for this overestimation, ensuring that the complete sum (all IMFs plus the residue) reconstructs the original signal exactly. This property is consistent with the mathematical definition of EEMD, where the reconstruction identity is strictly fulfilled only when both the IMFs and the residue are included and does not constitute an algorithm error but rather an inherent characteristic of the adaptive decomposition process.

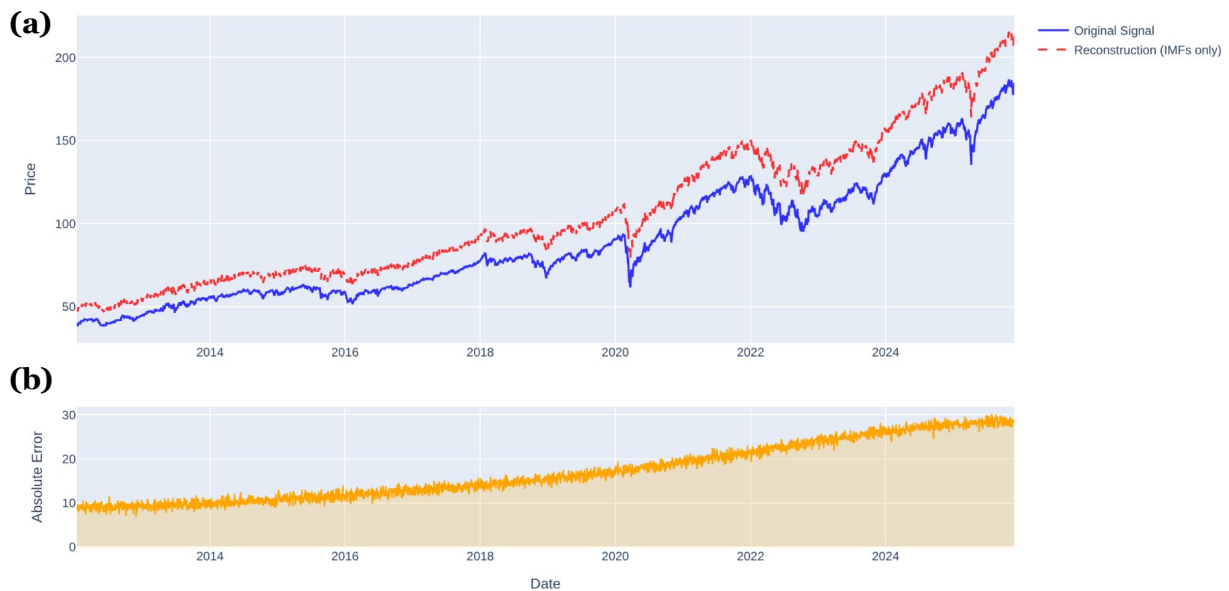

**Figure 2.** EEMD decomposition of the MSCI World index into 10 IMFs and residue. a) Com- parison between the original signal of the MSCI World index and its reconstruction using only the sum of the 10 IMFs extracted via EEMD, without including the residue. b) Absolute reconstruction error, calculated as the difference between the original signal and the sum of the IMFs.

**3.3 Graph transformation**

Graph transformations were performed via visibility and recurrence algorithms for all IMFs resulting from the decomposition. To visually illustrate the resulting structure of these transformations, Figure 3 shows the HVG graph obtained from IMF1, which represents the highest frequency component of the decomposition. The visualization reveals the characteristic topology of HVG graphs: a sparse structure where each node (corresponding to a temporal point of the series) connects only with those nodes that satisfy the horizontal visibility criterion. This structure reflects the high-frequency nature of IMF1, where rapid oscillations generate localized connectivity patterns that preserve the temporal information of the original series. The graphical representation allows us to appreciate how the visibility algorithm captures the geometric relationships between data points, transforming the one-dimensional temporal sequence into a two-dimensional network structure that maintains the structural properties of the modal component.

Specifically for the construction of recurrence graphs, optimal embedding parameters were determined independently for each IMF using the methods described in Section 2. The delay parameter $\tau$ was selected through mutual information analysis, identifying the first minimum of the mutual information function between the original series and its time-delayed version. The embedding dimension d was estimated using the false nearest neighbors (FNN) method, systematically increasing the dimension until the proportion of false neighbors fell below the 10% threshold.

The recurrence threshold $\varepsilon$ was established as the 10th percentile of the distribution of pairwise Euclidean distances between all embedding vectors, ensuring that the graph captures the most significant recurrent structures. The parameter selection results reveal considerable variability among the different IMFs: high-frequency IMFs (IMF1 to IMF4) require embedding dimensions of d = 4 and relatively small delays ($\tau$ between 2 and 6), while low-frequency IMFs (IMF8 to IMF10) and the residue use smaller dimensions (d = 2 or d = 3) and higher delays ($\tau$ between 16 and 41). This variability reflects the different temporal scales and dynamic properties inherent to each modal component. Table 3 presents the selected parameters for each IMF and the residue.

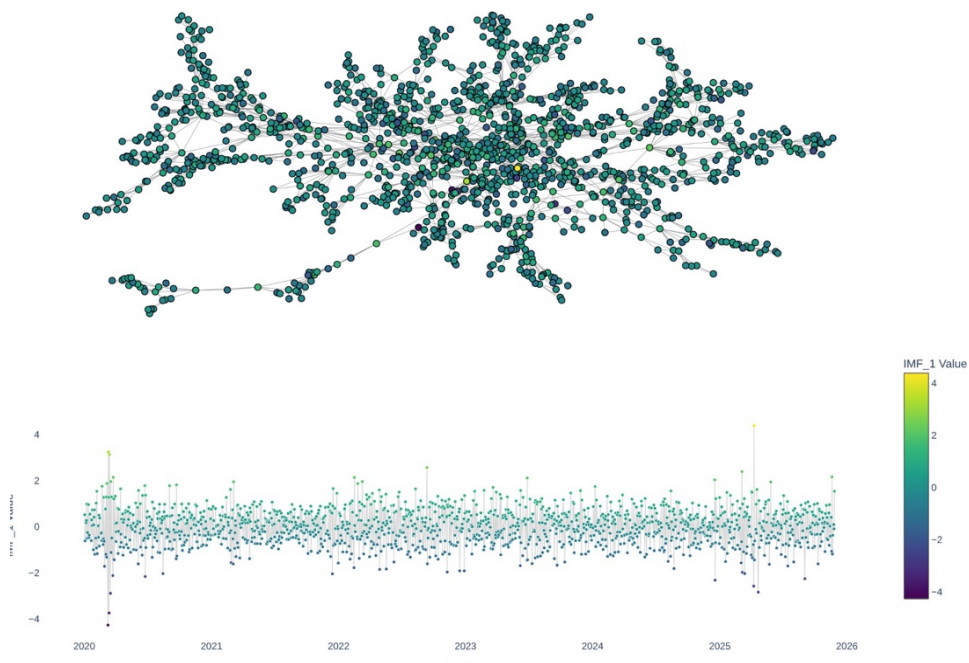

**Figure 3.** Horizontal visibility graph resulting from the transformation of IMF1 of the MSCI World index. Each node represents a temporal point of the series, and the connections reflect the horizontal visibility criterion.

**Table 3.** Embedding parameters and threshold ε by IMF

| Component | τ | d | ε |
|---|---|---|---|
| $IMF_1$ | 2 | 4 | 0.3620 |
| $IMF_2$ | 2 | 4 | 0.2251 |
| $IMF_3$ | 3 | 4 | 0.1829 |
| $IMF_4$ | 6 | 4 | 0.2123 |
| $IMF_5$ | 11 | 4 | 0.2481 |
| $IMF_6$ | 26 | 3 | 0.1721 |
| $IMF_7$ | 41 | 3 | 0.1324 |
| $IMF_8$ | 33 | 2 | 0.0720 |
| $IMF_9$ | 16 | 2 | 0.0636 |
| $IMF_{10}$ | 25 | 2 | 0.1576 |

**3.3.1 General structural properties. Connectivity and distance**

The comparative analysis of the fundamental structural properties of graphs generated through the three transformation methods (NVG, HVG, and recurrence) reveals significant differences in the topology of the resulting networks. These differences reflect the inherent characteristics of each transformation algorithm and its interaction with the dynamic properties of the different IMFs.

In terms of network size, the three methods generate graphs with a similar number of nodes, approximately 3490 nodes corresponding to the daily observations of the time series. However, recurrence graphs present slightly fewer nodes (between 3408 and 3484) due to the embedding process that requires delay and dimension parameters, which reduces the number of points available for graph construction. This difference is minimal and does not significantly affect the comparative analysis.

The most distinctive characteristic among transformation methods is connection density, measured through the number of links and graph density. HVG graphs exhibit an extremely sparse structure, with densities ranging between 0.00057 and 0.00114, corresponding to between 3489 and 6968 links. This low density reflects the restrictive nature of the horizontal visibility criterion, which only connects points when all intermediate points are below the minimum of the two considered points. In contrast, recurrence graphs present moderate and relatively stable densities, with values between 0.00264 and 0.00293, corresponding to between 16033 and 17753 links. This intermediate density arises from the distance threshold used in the construction of the recurrence matrix, which captures similarity patterns in the embedding space.

NVG graphs show the greatest variability in density, with values ranging from 0.00161 (for IMF1) to 0.70455 (for IMF10), representing a range between 9828 and 4289492 links. This extreme variability reflects how the natural visibility criterion, less restrictive than HVG, responds differently to the characteristics of each IMF. Higher-order IMFs, which capture long-term trends and low- frequency components, generate NVG graphs with extremely high densities, while high-frequency IMFs produce sparser structures. This property makes NVG graphs especially sensitive to the spectral characteristics of each modal component.

The global connectivity of graphs, measured through the number of connected components, reveals another fundamental difference among methods. Both HVG and NVG graphs are always fully connected (a single component), which facilitates message passing in GNN architectures and allows the analysis of global properties without the need for special techniques to handle disconnected components. In contrast, recurrence graphs present multiple disconnected components, with a number varying between 1 (for IMF9, residue, and IMF10) and 1024 (for IMF3). This fragmentation arises from the nature of the embedding space and the distance threshold used, where regions of phase space that do not share nearby neighborhoods give rise to separate components.

This property requires special strategies in GNN modeling, such as component-wise processing or pooling techniques that aggregate information from multiple components.

The average degree of nodes provides an additional measure of local connectivity structure. HVG graphs present very low average degrees, between 1.99 and 3.99, confirming their extremely sparse nature. Recurrence graphs show moderate and relatively stable average degrees, between 9.20 and 9.78, indicating consistent local connectivity regardless of IMF order. NVG graphs exhibit the greatest variability in average degree, from 5.63 (for IMF1) to 2458.16 (for IMF10), reflecting the dependence of graph density on the spectral characteristics of each IMF. This extreme variability in NVG graphs may pose challenges for learning consistent representations in GNN architectures, as different IMFs generate structures with radically different connectivity properties.

In summary, the three transformations generate graph structures with distinctive topological properties. On one hand, HVG graphs are extremely sparse and always connected, while recurrence graphs present moderate densities but multiple disconnected components. Finally, NVG graphs show highly variable densities that depend strongly on IMF order. These structural differences have important implications for the design of GNN architectures and the selection of appropriate modeling strategies for each type of transformation. Table 4 presents a summary of all structural properties obtained on the graphs after their transformation.

### 3.3.2 Clustering coefficients and local structure

The clustering coefficient, which quantifies the tendency of nodes to form triangles or local cliques, addresses the question of how cohesive local neighborhoods are, revealing notable contrasts among transformation methods.

**Table 4.** Summary of structural properties of transformed graphs

| Property | HVG | Recurrence | NVG |
|---|---|---|---|
| Density (range) | 0.00057 - 0.00114 | 0.00264 - 0.00293 | 0.00161 - 0.70455 |
| Number of links (range) | 3489 - 6968 | 16033 - 17753 | 9828 - 4289492 |
| Connected components | 1 (always) | 1 - 1024 | 1 (always) |
| Diameter (range) | 21 - 3489 | 20 - 1110 | 10 - 272 |
| Average degree (range) | 1.99 - 3.99 | 9.20 - 9.78 | 5.63 - 2458.16 |
| Average eccentricity (range) | 16.55 - 2617.0 | 13.99 - 862.25 | 8.06 - 266.70 |

A direct comparison of average ranges places NVG graphs at the top (0.5303-0.8310), followed by recurrence graphs (0.2636-0.6321), and finally HVG graphs (0.0-0.6594). This hierarchy reflects fundamental differences in how each transformation criterion structures local connections: the natural visibility criterion (NVG) generates denser neighborhoods, while the geometric constraints of the HVG algorithm produce sparser structures.

However, this general hierarchy conceals subtler patterns that depend on IMF order. In NVG graphs, high-frequency IMFs (IMF1 to IMF4) maintain coefficients between 0.6678 and 0.7547, but these values gradually decrease in the intermediate range (IMF5 to IMF9) until reaching 0.5303 for IMF9. IMF10, which captures the long-term trend, breaks this pattern with the highest value (0.8310), suggesting that low-frequency components generate highly cohesive local structures when the natural visibility criterion is applied. Standard deviations, which range between 0.1920 (IMF4) and 0.3332 (IMF9), reveal moderate variability within each graph. Medians, ranging from 0.6349 (IMF8) to 0.8966 (IMF10), confirm that most nodes in NVG graphs exhibit high clustering values.

The behavior of HVG graphs contrasts dramatically. Although high-frequency IMFs (IMF1 to IMF4) preserve some local cohesion (0.5314-0.6594) despite the constraints of the horizontal visibility criterion, this cohesion completely vanishes in low-frequency IMFs: IMF10 and the residue reach values of 0.0, indicating complete absence of clustering. This systematic degradation reflects how the geometric constraints of the HVG algorithm prevent triangle formation in extremely sparse structures. Variability, measured by standard deviations between 0.0 and 0.2893, is maximum in high-frequency IMFs. Medians, ranging from 0.0 to 0.6667, with values of 0.5 for most intermediate IMFs, reveal a bimodal distribution where many nodes completely lack clustering while others maintain moderate values.

Recurrence graphs display dynamics opposite to HVG graphs. While in HVG graphs local cohesion decreases with IMF order, in recurrence graphs the opposite occurs: the residue (0.6321), IMF10 (0.6133), and IMF9 (0.4696) exhibit the highest values, while high-frequency IMFs (IMF1 to IMF4) present the lowest (0.2636-0.3243). This inverse pattern suggests that the recurrence structure better captures local cohesion in low-frequency components, where more stable and recurrent dynamic patterns facilitate the formation of dense local communities. Considerable variability (standard deviations between 0.0502 for the residue and 0.3250 for IMF8) indicates heterogeneity in local structure. Medians, ranging between 0.3333 (IMF3 and IMF4) and 0.6429 (IMF10), reflect this variability according to IMF order.

Table 5 synthesizes these results, showing the ranges of average, median, and standard deviation values for each method.

The local structure of transformed graphs, characterized through clustering coefficients, reveals distinctive topological properties that have important implications for GNN modeling. NVG graphs, with their high clustering coefficients, present dense and cohesive local neighborhoods that facilitate localized message passing and can capture complex interaction patterns. HVG graphs, with their variable and frequently low clustering coefficients, especially for low-frequency IMFs, generate sparser local structures that may require different aggregation strategies in graph convolution layers. Recurrence graphs, with their intermediate clustering coefficients and inverse pattern according to IMF order, offer a local structure that reflects the dynamic properties of the underlying system, where low-frequency components show greater local cohesion due to the recurrent nature of their dynamic patterns.

**Table 5.** Clustering coefficients by transformation method and IMF order

| Method | Average coefficient (range) | Median coefficient (range) | Standard deviation (range) |
|---|---|---|---|
| HVG | 0.0 - 0.6594 | 0.0 - 0.6667 | 0.0 - 0.2893 |
| Recurrence | 0.2636 - 0.6321 | 0.3333 - 0.6429 | 0.0502 - 0.3250 |
| NVG | 0.5303 - 0.8310 | 0.6349 - 0.8966 | 0.1920 - 0.3332 |

**3.3.3 Centrality analysis**

Identifying which nodes play critical roles in the graph structure is fundamental for understanding how information flows and where influence is concentrated. For this purpose, we examine three centrality metrics that capture complementary aspects: betweenness centrality identifies strategic bridges, closeness centrality reveals nodes with rapid access to the rest of the network, and eigenvector centrality measures influence propagated through connections.

When analyzing which nodes act as critical bridges in information flow, marked contrasts emerge among transformation methods. NVG graphs show extremely low average values of between- ness centrality (between 0.0006 and 0.0058), suggesting that high density uniformly distributes the intermediation role, avoiding dependence on specific nodes as bridges. In contrast, HVG graphs exhibit dramatic variability: while high-frequency IMFs (IMF1) present average values of 0.0026, low-frequency IMFs (IMF10 and residue) reach 0.3333, with maximum values reaching up to 0.6539 for IMF6. This disparity arises because the extremely sparse structure of low- frequency IMFs concentrates information flow in few strategic nodes.

Recurrence graphs occupy an intermediate position, with average values between 0.0006 (IMF3) and 0.1094 (IMF10), where the presence of multiple disconnected components reduces the need for centralized intermediary nodes.

The ability to rapidly access the rest of the network, measured by closeness centrality, shows a distinct pattern. NVG graphs stand out with the highest values (range between 0.0680 for IMF9 and 0.2503 for IMF7), confirming that their high density allows most nodes to maintain short distances between each other. Both HVG and recurrence graphs present considerably lower values, although with subtle differences: in HVG graphs, values range between 0.0009 (IMF10 and residue) and 0.1019 (IMF1), while in recurrence graphs the range goes from 0.0028 (IMF10) to 0.0993 (IMF1). In both cases, high-frequency IMFs show greater accessibility due to their more compact structures and reduced diameters.

Eigenvector centrality, which evaluates influence based on the importance of neighbors, reveals important structural limitations. HVG and recurrence graphs share very low average values (between 0.0016-0.0058 and 0.0025-0.0058, respectively), although with differences in maximum values: HVG graphs reach up to 0.6186 for IMF6, while recurrence graphs rarely exceed 0.1876. This disparity in maxima reflects that, although the sparse structure of HVG graphs limits general influence propagation, some nodes can maintain significant local influence. NVG graphs present higher average values (between 0.0045 and 0.0159), with maximum values varying considerably from 0.0698 (IMF7) to 0.4336 (IMF2), suggesting that high density facilitates influence propagation, although in a more distributed manner in higher-order IMFs.

**Table 6.** Centrality measures by transformation method

| Metric | HVG (average range) | Recurrence (average range) | NVG (average range) |
|---|---|---|---|
| Betweenness | 0.0026 - 0.3333 | 0.0006 - 0.1094 | 0.0006 - 0.0058 |
| Closeness | 0.0009 - 0.1019 | 0.0028 - 0.0993 | 0.0680 - 0.2503 |
| Eigenvector | 0.0016 - 0.0058 | 0.0025 - 0.0058 | 0.0045 - 0.0159 |

Table 6 synthesizes the observed ranges for each metric. These structural differences have direct implications for modeling: NVG graphs facilitate efficient information propagation but with distributed influence, HVG graphs generate structures where certain nodes acquire critical roles as intermediaries (especially in low-frequency IMFs), while recurrence graphs, with generally low values across all measures, reflect more localized structural importance within fragmented components.

**3.3.4 Patterns according to IMF order**

The analysis of structural properties of transformed graphs reveals systematic patterns that vary according to IMF order, reflecting the different temporal scales and dynamic properties inherent to each modal component. These patterns provide valuable insights into how the spectral characteristics of IMFs translate into topological properties of the resulting graphs.

HVG graphs exhibit progressive structural degradation as IMF order increases. In high- frequency IMFs (IMF1 to IMF4), structures are relatively compact: diameters between 21 (IMF1) and 63 (IMF4), moderate densities (between 0.0011 and 0.0011), and high clustering coefficients (between 0.5314 and 0.6594). However, this initial cohesion systematically vanishes in higher- order IMFs. Diameters surge to 3489 for IMF10 and residue, while densities fall to 0.0006 and clustering coefficients reach zero. The underlying explanation lies in that the restrictive horizontal visibility criterion, although effective for capturing rapid oscillations, becomes increasingly limiting when applied to low-frequency components, producing extremely linear and sparse structures without local cohesion.

In contrast to HVG graphs, recurrence graphs show opposite dynamics: low-frequency IMFs develop notably more cohesive structural properties. High-frequency IMFs (IMF1 to IMF4) are characterized by multiple disconnected components (between 539 and 1024 components), relatively small diameters (between 20 and 34), and low clustering coefficients (between 0.2636 and 0.3243). The transition toward higher-order IMFs entails marked structural consolidation: the number of components reduces to 1 for IMF9, IMF10, and residue, diameters grow to 1110 for IMF9, and clustering coefficients increase significantly to 0.6321 for the residue. This evolution is due to low-frequency components, with more stable and recurrent dynamic patterns, generating more cohesive embedding structures where phase space points cluster more densely, favoring the formation of more compact local communities.

The behavior of NVG graphs differs substantially from the previous ones, showing non- monotonic variations in structural properties according to IMF order. High-frequency IMFs (IMF1 to IMF4) are distinguished by relatively low densities (between 0.0016 and 0.0052), small diameters (between 10 and 14), and high clustering coefficients (between 0.6678 and 0.7547). In the intermediate range (IMF5 to IMF9), density progressively increases to 0.1703 for IMF9, diameters grow moderately to 127 for IMF9, while clustering coefficients gradually decrease to 0.5303 for IMF9. IMF10 breaks this trend with extreme properties: density of 0.7045, diameter of 272, and the highest clustering coefficient (0.8310). This non-monotonic variability arises because the natural visibility criterion responds heterogeneously to the spectral characteristics of each IMF, generating highly dense structures in low-frequency components where most points satisfy the visibility criterion.

These systematic patterns according to IMF order have important implications for GNN architecture design. High-frequency IMFs, with their more compact and cohesive structures in HVG and NVG graphs, can benefit from standard graph convolution layers that leverage local cohesion. Low-frequency IMFs require adaptive strategies: in HVG graphs, where structure is extremely sparse, attention techniques or adaptive pooling may be needed; in recurrence graphs, where structure is more cohesive, standard layers can be effective; and in NVG graphs, where density is extremely high, sampling or pooling techniques may be required to handle computational complexity.

**3.3.5 Implications for GNN modelling**

The structural properties identified in transformed graphs have important implications for the ap- plication of graph neural networks. NVG graphs, with their high density and very short-distance structure, provide a rich environment for message passing in GNNs, allowing information to prop- agate efficiently through the network. However, high density can also increase computational complexity and potentially introduce noise in learned representations.

HVG graphs take a different approach. Their sparser, more uniform structure yields a leaner representation that reduces computational overhead. Still, the wide range of distances observed particularly in low-frequency IMFs can complicate the learning of stable representations. One clear benefit is that all HVG graphs remain connected, enabling global message passing without resorting to specialized techniques for handling isolated components.

Recurrence graphs introduce another layer of complexity. Most IMFs contain numerous dis- connected components, forcing GNN designers to adopt component-wise processing or pooling methods that merge information across components. Despite this challenge, recurrence graphs encode core dynamic features of the underlying system, making them useful for prediction tasks.

Taken together, these three methods complement each other. A hybrid approach that integrates multiple graph representations could exploit the advantages of each while compensating for their weaknesses. This direction opens up promising avenues for developing multi-view GNN architectures tailored to financial time series decomposed through EMD.

## 4. Conclusions and future

### 4.1 Conclusions

As a result of this study, a methodological framework has been developed that combines EMD with graph transformation techniques to analyze financial time series, specifically applied to the MSCI World index. The results obtained show that this integration of complementary techniques enables a deeper understanding of the complex, nonlinear, and non-stationary dynamics that characterize financial markets.

Statistical validation using the ADF, KPSS, and BDS tests confirmed that the MSCI World index satisfies the necessary conditions to apply EMD effectively, establishing a solid foundation for subsequent analysis. Decomposition using EEMD produced 10 IMFs plus a monotonic residue, each capturing different temporal scales of market dynamics. Particularly relevant is the finding that higher-order IMFs concentrate most of the energy and variance of the system, where IMF10 contains 88.47% of the total energy, representing long-term trends, while lower-order IMFs capture high-frequency oscillations and short-term noise.

The transformation of these IMFs into graph representations through three complementary methods reveals distinctive topological properties that reflect different aspects of the underlying structure of financial data. These properties can be summarized as follows:

- HVG graphs generate extremely sparse structures with low density and average degree while maintaining full connectivity. However, they exhibit high variability in distances for low-frequency IMFs, which complicates the learning of consistent representations.

- Recurrence networks show moderate and stable densities, but frequently present multiple disconnected components, especially for high-frequency IMFs. This fragmentation requires specialized neural network architecture design strategies for effective component processing.

- NVG graphs exhibit the greatest variability in density, ranging from sparse structures for high-frequency IMFs to extremely dense graphs for low-frequency IMFs. This high density, while capturing rich relationships, significantly increases computational complexity.

A key finding obtained from the analysis is that the structural properties of the graphs vary systematically with the order of the IMFs, but each transformation method exhibits distinctive patterns. HVGs show progressive structural degradation, recurrence networks present greater cohesion for low-frequency IMFs, and NVGs exhibit non-monotonic variations. This diversity suggests that no single graph representation method is optimal for all IMF components.

The results of this study provide crucial insights for the design of GNN architectures applied to financial time series forecasting. The identified topological properties (extreme sparsity in HVGs, disconnected components in recurrence networks, and high density in NVGs) pose specific modeling challenges that must be addressed through specialized architectural strategies. Hybrid approaches are required that can leverage the complementary strengths of multiple graph representations simultaneously.

In conclusion, this work establishes that the integration of EMD with graph analysis constitutes a powerful tool for unraveling the multilevel structure of financial time series. The discovered topological patterns not only deepen our understanding of market dynamics, but also inform the development of more sophisticated GNN-based forecasting systems. The proposed methodology is generalizable to other financial indices and asset classes, opening promising avenues for future research in quantitative financial market analysis.

**4.2 Future lines**

The results and limitations identified in this work open multiple promising directions for future research. A natural extension of this work is the development and implementation of specialized GNN architectures that exploit the topological properties identified in the graph representations of the IMFs. Given that each graph transformation method captures complementary aspects of market dynamics, it is proposed to develop multi-view GNN architectures that simultaneously process multiple graph representations (17; 18). These architectures could employ adaptive attention mechanisms to dynamically weight the contribution of each view according to the prediction time horizon and market conditions (19).

Recent literature demonstrates the effectiveness of combining temporal encoders with graph convolution layers to capture both temporal dependencies and spatial relationships among assets (17; 18; 20). For the IMFs extracted in this study, it is recommended to develop models that dynamically adapt the graph structure over time, allowing relationships among components to evolve according to market states. Additionally, the recurrence networks used in this work suggest the presence of higher-order relationships among IMFs. Architectures based on adaptive hypergraphs (19; 21) could capture these multi-component associations more effectively than traditional graphs. Specifically, it is proposed to investigate hypergraph attention networks in hyperbolic spaces to model hierarchical relationships among IMFs of different temporal scales (21), heterogeneous graphs that incorporate multiple types of relationships among components (20), and dual attention mechanisms that operate at both the node level and the relationship-type level to refine information propagation (18).

Scalability is fundamental for the practical application of GNNs in financial markets, where the number of assets, data update frequency, and relationship complexity can grow exponentially. Handling graphs in distributed environments represents a critical research line. It is proposed to investigate the adaptation of modern distributed graph processing frameworks to the specific context of financial time series. Current distributed systems employ graph partitioning strategies, multi-level parallelism, and communication optimizations to scale GNN training (24). In order to achive this goal, it is recommended to evaluate partitioning strategies that respect the temporal structure of the IMFs, minimize communication between workers, implement data parallelism to process multiple temporal windows simultaneously, and develop caching schemes and pre- computation of embeddings for stable long-term IMFs.

Financial markets generate data continuously, requiring incremental updating of graph representations. It is proposed to investigate incremental EMD update algorithms that allow new observations to be incorporated without recomputing the entire decomposition, GNN architectures for dynamic graphs that support continuous updates of structure and attributes (22; 23), and real-time inference systems that balance latency and accuracy through distributed computation. Dense NVG graphs and recurrent networks with multiple components pose challenges in terms of memory footprint and communication bandwidth in distributed environments. Future lines include graph compression techniques that preserve topological properties essential for learning, adaptive neighborhood sampling strategies that reduce complexity without degrading prediction quality, and asynchronous communication protocols to minimize global synchronizations during distributed training.

Extending the analysis from a single index to complete portfolios of global assets requires robust distributed processing capabilities. It is proposed to develop federated architectures that enable training models on data from multiple markets while preserving privacy, implement transfer learning systems that leverage knowledge from one market to improve predictions in others, and design multi-index aggregation strategies that capture dependencies between regional and global markets. Other research directions include:

- Apply the proposed methodological framework to different financial instruments to evaluate the generalization of the identified topological patterns,

- Research how the topological properties of graphs derived from IMFs vary across different market states and exploit these patterns for early detection of regime changes,

- Develop interpretation techniques for GNN models that enable the identification of which IMF components and which graph structures contribute most to predictions,

- Investigate adaptive methods to automatically select optimal EEMD parameters according to the specific characteristics of each time series,

- Asses other types of time series-to-graph transformations and explore the incorporation of non-traditional data sources (i.e news, social media data, sentiment analysis, etc) into graph representations to enrich the predictive power of GNN models.

Hence the proposed future directions range from the development of specialized GNN architectures that exploit the identified topological properties to the implementation of scalable distributed systems capable of processing dynamic graphs in real time. The convergence of signal decomposition techniques, graph theory, deep learning, and distributed computing promises significant advances in financial time series forecasting, with potential impact both in academic and practical trading and risk management applications.